\documentclass{article}
\usepackage{aaai}

\usepackage{amsmath}
\usepackage{amssymb}
\usepackage{theorem}
\usepackage{moreverb}

\newcommand{\specialchar}[1]{\ensuremath{{\cal #1}}}

\newcommand{\ttt}{\ensuremath{\overline{t}}\space}	

\newcommand{\xx}{\ensuremath{\overline{x}}\space}	

\newcommand{\XX}{\ensuremath{\overline{X}}\space}	

\newenvironment{notation}[1][notation]{\par\noindent\textsc{#1}\newline\tt}{}
\newenvironment{notation*}{\begin{notation}[]}{\end{notation}}

\newcommand{\ra}{\rightarrow}
\newcommand{\la}{\leftarrow}
\newcommand{\lra}{\leftrightarrow}

\title{Problem solving in ID-logic with aggregates: \\some experiments}
\author{Bert Van Nuffelen \and Marc Denecker\\
K.U.Leuven, Department of Computer Science\\
Celestijnenlaan 200A, 3001 Heverlee, Belgium\\
\{marcd,bertv\}@cs.kuleuven.ac.be\\
}

\nocopyright

\newcommand{\Dlogic}{ID-logic}
\newcommand{\defp}[1]{Defined(#1)}
\newcommand{\rules}[1]{Rules(#1)}

\newcommand{\state}{\specialchar{S}}

\newcommand{\CS}{\specialchar{CS}}

\begin{document}

\maketitle

\begin{abstract}

The goal of the LP+ project at the K.U.Leuven is to design an
expressive logic, suitable for declarative knowledge representation,
and to develop intelligent systems based on Logic Programming
technology for solving computational problems using the declarative
specifications. The ID-logic is an integration of typed classical
logic and a definition logic. Different abductive solvers for this
language are being developed. This paper is a report of the integration 
of high order aggregates into ID-logic and the consequences on the 
solver SLDNFA.

\end{abstract}

\section{Introduction}

The goal of computational logic is to design logics for {\em knowledge
  representation} and to develop algorithms to solve computational
problems using the {\em declarative specifications}.  In principle,
the declarative knowledge representation methodology in logic is based
on a simple idea. To describe his knowledge, an expert designs
the ontology of his problem domain: he defines the relevant types of
objects, and the relevant relations and functions between these
objects and chooses a logical alphabet to name them.  In the next
phase, the expert uses this alphabet to express his knowledge by a
set of logical sentences that are {\em true} statements on the problem
domain.

In the choice of the alphabet, the expert may be led by different and
often non-compatible quality criteria: naturality of the
representation, expressivity of the logic, efficiency of the
representation, etc.. However, if the goal is to obtain a clear
declarative representation, the expert will choose the alphabet as
close as possible to what he views as the relevant objects, concepts
and relationships in the problem domain. Alphabets matching more
closely the natural ontology lead to more natural representations.

Choosing the alphabet in accordance with the natural ontology of the
problem domain imposes high requirements both on the expressivity and
on the problem solving capabilities of the used logic. With respect to
the expressivity, higher order aggregates easily show up in
declarative representations of practical problem domains. For example
in the domain of university lecture scheduling one of the
constraints will be that if a lecture $l$ of course $c$ takes place in
a room $r$ with a capacity of $rc$ students, then the number $n$ of
students enrolled for course $c$ must be less than $rc$.  In a
suitable logic, the constraint can be represented using the
cardinality aggregate:
\[ \begin{array}{l}
  \forall\ c,l,r,rc,n.\ room\_of\_lecture(l,r) \land capacity(r,rc)\\
\land course(l,c) \land Card(\{st | enrolled(st,c)\}, n)  \ra rc \geq n
\end{array} \]
Other frequently occurring aggregates are \textsl{summation},
\textsl{minimum}, \textsl{maximum}, etc \dots  Therefore aggregates
are extensively studied in logic programming and deductive
databases \cite{KempStuckey91,VanGelder92,Ross92}.

A natural choice of the alphabet also poses requirements on the level
of problem solving capabilities of the used logic. Illustrated in the
domain of university lecture scheduling, some basic concepts are {\em
  lectures}, {\em time slots}, {\em rooms}.  Basic relations between
these concepts describe when and where lectures take place. The
natural choice to represent these relations are by (typed) predicates
(e.g. $time\_of\_lecture/2$ and $room\_of\_lecture/2$).  Now, consider
the task of computing a schedule satisfying certain data and
constraints (lectures to be given, rooms and lecturers available, no
overlap, etc\dots). This is a satisfiability problem or an abductive
problem; tables of these predicates must be computed that satisfy the
data and constraints imposed on correct schedules.  Note that at this
point, we have only chosen the alphabet; we have not formalized one
single iota of knowledge in a logical formula. This shows that often,
satisfiability checking or abduction is a natural companion and a
natural consequence of applying the declarative knowledge
representation methodology.

At the K.U.Leuven, the aim of the LP+ project is to use both semantical and implementational techniques of logic programming to develop a logic suitable for declarative knowledge representation and to implement efficient problem solvers for such a logic.
In this paper, we study satisfiability problems in the context of a logic with higher order aggregates.
The used logic is an extension of inductive definition logic (ID-logic) \cite{Denecker00,Denecker98c,Denecker95a} developed within the project.
ID-logic is a conservative extension of classical logic with a generalized notion of non monotone inductive definitions.
At the same time, the logic can be seen as a natural generalization of Abductive Logic Programming \cite{Kakas93a} and Open Logic Programming \cite{Denecker95a}.

The technology that is used so far for reasoning and problem solving
in ID-logic is based on an integration of techniques of abductive
logic programming \cite{Kakas93a} and constraint logic programming
techniques \cite{JaffarMaher94}. The first published presentation of
such a procedure was in \cite{Kakas95a,Kakas99a}.
\cite{VanNuffelen99} describes a similar integration of the abductive
resolution method SLDNFA \cite{Denecker98z} with CLP techniques and
presents some experiments in the context of ID-logic specifications of
some typical CLP-problems (such as N-queens), and of scheduling and
planning.

In this paper, we define an extension of ID-logic with aggregates, sketch an extension of the solver of \cite{VanNuffelen99} to reason on aggregates and describe two computational experiments with the system for solving a sea-battle puzzle and a scheduling problem.
At the logical level, our work is based on Van Gelders's work \cite{VanGelder92} on aggregates in the context of well-founded semantics.
On the other hand the implementation and experiments extend previous work described in \cite{Denecker97b,Tractebel}.
To the best of our knowledge, this paper is the first report on solving satisfiability and abductive problems in the context of (an extension of) classical logic with higher order aggregates.
All the studies concerning aggregation we encountered, were in the context of querying systems.

  
Because an abductive system for ID-logic cannot be complete (computing
satisfiability of a classical first order theory is a undecidable
subproblem of the computation of an abductive solution of an ID-logic
theory.), the current implementation is only valid for a restricted
class of problems.  But the experiments show the feasibility of
reasoning on a useful class of specifications with aggregates. 

\section{ID-logic extended with aggregates}

\subsection{ID-logic}
\label{SIDlogic}

As mentioned, ID-logic is an extension of classical first order logic
(FOL) with inductive definitions. The logic builds upon the earliest
ideas on the declarative semantics of logic programs with negation as
failure. The view of a logic program as a definition of its predicates
is underlying both the least model semantics of van Emden and Kowalski
\cite{vanEmden76} and Clark's completion semantics \cite{Clark78}.
This idea was further explored in \cite{Denecker98a}, where it was
argued that the well-founded semantics for logic programming
implements a generalized principle of non-monotone induction. A
discussion of this is out of the scope of this paper.

An ID-logic theory $T$ consists of a set of {\em definitions} and a
set of classical logic sentences. A definition is an expression that
{\em defines} a subset of predicates in terms of the other predicates.
Formally, a definition $D$ is a pair of a set $\defp{D}$ of predicates
and a set $\rules{D}$ of rules that exhaustively enumerate the cases
in which the predicates of $\defp{D}$ are true. A rule is of the form:
\[ p(\ttt) \la F\]
where $p \in \defp{D}$ and $F$ an arbitrary first order formula. The
predicates in $\defp{D}$ are called {\em defined} by $D$, the others
(not in $\defp{D}$) are called {\em open} in $D$.

The semantics of ID-logic integrates classical logic semantics and
well-founded semantics. An interpretation $M$ is a model of a
definition $D$ iff it is total (i.e. 2-valued) and the unique
well-founded model of $D$ extending some interpretation $M_o$ of the
functor and open predicate symbols of $D$. An interpretation $M$ is a
model of an \Dlogic\ theory $T$ iff it is a total model of its
classical logic sentences and of its definitions.

Logical entailment is defined as usual: $T \models F$ iff $F$ is true
in all models of $T$.

ID-logic generalizes not only classical logic but also abductive logic
programming \cite{Kakas93a} and open logic programming
\cite{Denecker95a}. An abductive logic framework, consisting of a set
of abducible predicates, a set of rules and a set of FOL {\em
  constraints} can be embedded in ID-logic as the theory consisting of
the FOL constraints and one definition defining all non-abducible
predicates. Formally, ID-logic extends ALP by allowing multiple
definitions and generalized syntax. However, it can be shown that it
is always possible to transform a set of definitions into one single
definition. As will appear in the next section, current problem
solvers for ID-logic are based on the technology of abductive logic
programming.

In the current version of the SLDNFA-system a simple many-sorted version of ID-logic is used. 
A type inference system checks and completes a partial set of user-defined type declarations: this produces a small overhead compared to the benefits it brings: e.g. error checking, disambiguating expressions, etc. For more details, we refer to \cite{DeMot99}. 
In the implementation, a Prolog-like style using capitals for
variables, and ``,'', resp.  ``;'' for conjunction, resp. disjunction is used.
For example a definition is represented as
{\small
\begin{verbatimtab}[3]
uncle(X,Y)<- ( exists(Z): parent(Y,Z),brother(X,Z); 
				   exists(A): aunt(A,Y),married(A,X) ).
							
aunt(X,Y) <- ( exists(Z): parent(Y,Z),sister(X,Z); 
               exists(A): uncle(A,Y),married(A,X)).
\end{verbatimtab}
}
\noindent This  definition defines the two predicates \textsl{uncle} and \textsl{aunt} simultaneously. The other are open predicates.

FOL axioms are represented in the system in the same style but are prefixed by the key-word {\tt fol}.
Examples are:
\begin{verbatimtab}[3]
fol forall(X,Y): 
	 uncle(X,Y), age(X,AgeX), ageY(Y,AgeY) 
	 => AgeX > AgeY.
fol aunt(mary,bob).
\end{verbatimtab}

{\bf A special case.} The definitions that appear in the experiments
in this paper are of a simple kind; they do not contain recursion. The
models of a definition without recursion are exactly the models of the
completion of the definition \cite{Clark78}.  Below, completed
definitions of predicates will be denoted: \[\forall(p(\overline{X})
\lra B_p[\overline{X}])\]
We call from now on $B_p[\XX]$ the \textsl{completion} of $p$.

\subsection{Aggregates}\label{semantics}
As mentioned in the introduction, aggregates have been studied in
logic programming and deductive databases.
\cite{KempStuckey91,VanGelder92,Ross92} proposed extensions of logic
programming with aggregates and showed how  aggregate expressions
can be transformed and reduced to recursive logic programs using
certain schemata. These transformations define at the same time the
declarative and procedural semantics for these extensions.  The papers
cover actually a subclass of ID-logic theories, because they don't
have the notion of open predicates.  In the context of ID-logic, where
we deal with open predicates, the proposed schemata (inductive
definitions) can be used to define the declarative semantics without
extending ID-logic, but the current implementation of the solver
cannot cope with the reduced programs due to their highly recursive
nature. If the current implementation would be extended with a proper
notion of tabling these definitions probably could be computed.
Therefore we take another approach by introducing new language primitives, which allows us to use aggregation without the computation of recursive definitions.

We define first two basic concepts of our aggregate expressions.
\begin{itemize}
\item  A \textbf{Set expression} is an expression of the form:
\[\{ \xx | F[\xx] \} \]
where \xx is a tuple of variables and $F[\xx]$ a first order formula.
It denotes the set of tuples $\xx$ that satisfy $F[\xx]$.  Three sorts
of variables are distinguished in a set expression: local variables
are quantified inside $F$; their scope is the quantifier. Parameter
variables are the elements of $\xx$; their scope is the set
expression. The other variables are free variables.

In the following example the set expression denotes for a person \texttt{Y} the set of all aunts of \texttt{Y} which are older than 50.
{ \small
\begin{verbatimtab}[3]
set([X],(exists(AgeX): 
			aunt(X,Y),age(X,AgeX), AgeX >50)))
\end{verbatimtab}
}
\item A \textbf{function expression} is of the form:
\[ \lambda \xx. y\ where\ F[\xx,y] \]
where \xx is a tuple of variables, $y$ is a variable and $F[\xx,y]$ is
a first order formula expressing a functional relation between $y$ and
$\xx$. I.e. it satisfies the constraint:
\[ \forall \xx \exists! y. F[\xx,y]\]
This expression denotes the anonymous function mapping a tuple \xx to the object
$y$ for which $F[\xx,y]$ is satisfied.  As above, we distinguish
between local variables, parameter variables and free variables of a
function expression.

A pure arithmetical function 
\[ \lambda \xx. t[\xx] \]
where $t[\xx]$ is an arithmetical term, is a shorthand for
\[ \lambda \xx. y\ where\ y=t[\xx] \]
The next example represents the function which maps a person X to his age.
{ \small 
\begin{verbatimtab}
lambda([X] , Y where (exists(Z) : age(X,Z), Y=Z))
\end{verbatimtab}
}
\end{itemize}

Set and function expressions are allowed to appear only in aggregate
expressions. The following aggregate primitives have been implemented:

\begin{itemize}
\item minimum: $minimum(\Psi,n)$ means that $n$ is the minimal element
  of the set denoted by $\Psi$.  In the current implementation, the
  set expression must be of type integer.
\item cardinality:  $card(\Psi,n)$ means that the set represented by $\Psi$ has $n$ elements. 
\item summation: $sum/3$ has as arguments an n-ary set expression, an
  n-ary function expression of a number type, and a number as
  argument. $sum(\Psi,f,s)$ means that:
\[ s=\sum_{\xx \in \Psi}f(\xx)\]
\item product: $product/3$ is analogous to $sum/3$ but expresses the
  product of a function over a set.
\end{itemize}

\subsection{Semantics}

In this section, we briefly explore how the semantics of ID-logic with
aggregates can be defined. The semantics of ID-logic can be extended
using the same transformational approach proposed by Van Gelder in
\cite{VanGelder92}. In this approach, aggregate expressions are
transformed to recursive logic programs under well-founded semantics.
Because the models of an ID-logic theory are well-founded models, the
same approach applies.

In the following example we illustrate the approach of
\cite{VanGelder92} in the case of the minimum-aggregate and show how
it can be transformed into a definition.  We illustrate the
transformation in the context of open predicates.  Consider the
following theory $T$ where $a$ is an open predicate ranging over
integers:
\[
T = \{ min({\{X|a(X)\}},4)  \}
\]
This theory expresses that the minimum of the argument of the predicate $a$ should be 4.
Applying the transformation scheme in \cite{VanGelder92}, we obtain: {
  \small
\begin{verbatimtab}[3]
        fol mina(4).

        mina(X)    <- a(X), not bettera(X).
        bettera(X) <- a(Y), Y<X.
\end{verbatimtab} 
} In this simple case, we obtain a non-recursive definition for the
new predicates $mina$ and $bettera$. Abductive solvers can be used to
successfully reason on these theories. For example
\begin{itemize}
\item The query\footnote{Note that in contrast to Logic Programming conventions, queries are not denoted by denials.}$true$ will lead to a successful derivation generating
  the abductive answer $\Delta = \{a(4)\}$. To solve the FOL
  constraint $mina(4)$, a(4) must be abduced.
\item Likewise, the query $a(6)$ will succeed with abductive answer
  $\Delta = \{a(4),a(6)\}$.
\item The query $a(1)$ will fail. During the derivation, the solver
  will abduce $a(1)$ and will attempt to solve $mina(4)$. This is
  impossible because $bettera(4)$ can be derived.
\end{itemize}

In the case of summation and cardinality, application of the schemata
results in highly recursive logic programs. As shown in
\cite{VanGelder92}, the transformational approach poses no problem at
the level of semantics; however SLDNFA loops on such programs.  For
this reason, we implemented aggregates in a different (more efficient)
way (see further on\ref{implementation}).

\section{An abductive problem solver}
\label{SecAbduction}
From here on, we will assume that an ID-logic theory $T$ contains only
one definition defining a number of predicates simultaneously. Recall
from previous section \ref{SIDlogic} that it is always possible to transform a
set of definitions into this form.

Given an ID-logic theory $T$ with definition $D$, an abductive problem
for a given query $Q$ consists of computing a set $\Delta$ of
definitions of ground atoms for each open predicate of $T$ and an
answer substitution $\theta$ such that $D + \Delta$ is consistent and
entails all FOL axioms in $T$ and $\forall(\theta(Q))$.  An abductive
procedure computes tables for the open predicates that can be extended
in a unique way to a well-founded model of the definition and a model
of the FOL axioms and of the query. This way, an abductive answer can
be seen as a compact space-efficient representation of a model.

SLDNFA \cite{Denecker92d,Denecker98z} is an abductive procedure for
normal logic programs. \cite{VanNuffelen99} describes an extension of
this procedure to deal with FOL axioms and generalized rules and
queries, and describes an integration of this procedure with
constraint solvers. This integration is in a similar spirit as ACLP
\cite{Kakas95a,Kakas99a}. Due to lack of space, we can only give the
head-lines of this procedure. For more detailed description, we refer
to \cite{VanNuffelen99,VanNuffelen00b}.

A derivation for a query $Q$
can be understood as a rewriting process of {\em states $\state$},
i.e. tuples $(\Theta,\Delta,\CS)$ of a set $\Theta$ of FOL formulas and
denials, a set $\Delta$ of abduced open atoms and a set $\CS$ of CLP expressions,
called the constraint store. A denial is a formula of the form
$\forall \overline{X}. \la F[{\overline{X},\overline{Y}}]$, where
$\la$ denotes negation.  Denials are the only formulas that may
contain universal quantifiers.  Free variables in FOL formulas and
denials represent objects of yet unknown identity.

The derivation starts with the initial state
$(\Theta,\emptyset,\emptyset)$ where $\Theta$ consists of $Q$ and the
set of FOL axioms in $T$. The rewriting process proceeds by selecting
an atom in a formula $F$ from $\Theta$ and computing a new state
depending on the sort of atom by applying the right rule. E.g. if $F$
is an open atom, the atom is abduced; an atom interpreted in a CLP
domain is added to the constraint store; defined predicates are
substituted by their completion and the resulting formula is then
simplified. Disjunctive goals are dealt with selecting alternative
disjuncts using backtracking.  Consistency of denials with a selected
open atom is checked by matching this open atom against each abduced
open atom in $\Delta$. If $fail$ is derived, the computation
backtracks.  The computation ends in three possible ways:
\begin{itemize}
\item with a \textsl{floundering} error condition when universally quantified
  variables appear in the selected atom in a denial.
\item with \textsl{failure}, if no solution is derived;
\item with a \textsl{successful} derivation if a state $\state$ is
  derived where \CS\ is a consistent constraint store, $\Delta$ a set
  of ground open atoms and $\Theta$ consists purely of denials that
  have been checked to be consistent with $\Delta$.
\end{itemize} 

\subsection{Extending the implementation for aggregates}
\label{implementation}

We extended SLDNFA in a heuristic manner to reason on aggregates.
If an aggregate expression is selected during the derivation, the set expression is rewritten using the completion of the defined predicates and the table of abduced atoms.
This process leads to a big disjunction enumerating potential values occurring in the set together with a CLP-constraint formula describing the logical conditions under which the potential value effectively belongs to the set.

When during the evaluation of a set expression open predicates are encountered, this partially evaluated expression is remembered and each time an atom is abduced later on, the procedure will check if it supports a new potential element of the set. Hence, abduction leads to new disjuncts in the set description.  

The unfolded disjunction of potential elements and associated constraints can be used then to compute the value of the aggregate expressions. For example, assume that for some cardinality expression $Card(\{ \xx| F[\xx] \}, N)$, $F[\xx]$ could be reduced to a disjunction \[\xx=\overline{v_1}\land C_1 \lor .. \lor \xx=\overline{v_n}\land C_n\] in which $v_i$ are distinct potential values and $C_i$ is the associated constraint of $v_i$.  
In this case the value of $N$ can be simply defined by the boolean sum  \[N = B_1 + .. + B_n \] where $B_i$ is a boolean variable defined by the constraint $B_i \iff C_i$. 
These kind of constraints are known as reified constraints \cite{SicstusCLPFD}.
The above sum can be efficient computed using specialized library constraints of the finite domain constraint solver.

The same principle can be applied in the case of minimum, maximum and summation and product.
For example, in the case of summation, each time a new potential value $v_i$ is derived for the set, the function value $f_i$ of this new potential element must be computed and the sum of the expression is computed as the sum of constraint expressions $B_i\times f_i$.

As can be seen above, the current implementation is strongly focussed on constructing a finite domain constraint store. 
This restricts, at this moment,  the set expressions to have the property that all the variables which value is unknown during the unfolding of the set expression and which have influence on the membership of a value in the set should be finite domain variables.
However this give us still the ability to reason on a large group of applications. 

Another restriction stems from the fact that the procedure will only observe the state and maintain a complete disjunction w.r.t. $\Delta$ during the evaluation of aggregate expressions.
The evaluation will not procedure new abduced atoms.
In general, it is easy to find applications axioms containing aggregate expressions should lead to new abductions.
For example, consider the following theory.
  {\small
\begin{verbatimtab}[3]
		fol Card(set([X],a(X)),3).
\end{verbatimtab} 
} 
\noindent To evaluate the query $true$ with respect to this theory, it is necessary to abduce 3 $a$-atoms. The current implementation fails on this query. 

It is a topic of future research to extend the current solver to deal with a broader class of problems. However, for an important class of practical applications, the solver works already fine. Namely when the open predicates appearing in the set expressions represent functions on some finite domain. In this case the solver will ultimately compute a complete table of abduced atoms of these predicates; consequently, the disjunctive representation of the set expressions will be complete as well.  As illustrated in the experiments, many problems satisfy this condition.


\subsection{Optimization functions}
Often, the expert is not interested in an arbitrary solution of an
abductive problem but in an optimal solution along some optimality
criterion. 
In general it is an intractable problem.
But in the context of Constraint Programming, one often recurs to the following pragmatic solution.
Given a constraint program and an optimization function, an initial solution is computed and the value of the optimization function for this solution is recorded; then the system backtracks and tries to find other solutions; the value of the optimization function of the best solution so far, is used to prune the search.
If the search stops, then it stops with an optimal solution; otherwise, the user may stop the system and extract the best solution so far.
In a lot of cases this is satisfying for the user.


We extended the abductive solver with a similar facility. Together
with the query, the user can specify an optimization function to be
minimized or maximized. This is done by specifying either
\texttt{minimize(V)} and \texttt{maximize(V)}, in which V is an
expression which should be minimized or maximized.  In practice,
because the search can take a long time (or does not end), the best
solution that can be computed within a given time is returned.

\section{Experiments}
We present two experiments: the first one is a solitaire puzzle based on the well-known battleship game.  The second one is about scheduling the maintenances of units in  power plants.  The experimental results are obtained using an implementation of SLDNFA as a meta program on Sicstus Prolog 3.7 on a Sun Solaris machine.  

\subsection{The battleship puzzle}
The objective of this puzzle \cite{battleship} is to find the 
locations of 10 ships hidden on a 10 by 10 board.  There are ships
from different sizes: one battleship, two cruisers, three destroyers
and four submarines.  The ships can be placed everywhere on the grid
either horizontally either vertically oriented.  They are not allowed
to touch each other, therefore a ship is always surrounded by water
(or the border of the grid).  The data consist of a given set of known
locations of boat pieces or water and the number of boat pieces on
each row and column.

The formalization of this puzzle in ID-logic starts with the choice of
the alphabet.
The central concept in this puzzle is the location of a
ship.  There are two options: either one represents the location by
one (X,Y) coordinate (e.g. the left upper one), the length and the
orientation of the ship; or one defines the location by means of
the locations of the different parts of a ship.  We have chosen 
the last option.
 
Let us define the battle fleet by the following facts\footnote{In the specifications we omit the type information}:
{\small
\begin{verbatimtab}[3]
ship(S) <- S in 1..10.

ship_type(1,battleship)<- true.          
ship_type(S,cruiser)   <- S in 2..3.     
ship_type(S,destroyer) <- S in 4..6.     
ship_type(S,submarine) <- S in 7..10.    

length(battleship,4)<-true.
length(cruiser,3)   <-true.
length(destroyer,2) <-true.
length(submarine,1) <-true.

ship_length(S,L) <- ship_type(S,Type), length(Type,L).
\end{verbatimtab}
}
\noindent \texttt{ship\_length} is an auxiliary definition which defines for a particular ship its length.
Then depending on the type, ships consist of different number of parts, each connected to a location.
We represent this by the open predicate \texttt{ship(S,P,X,Y)}, which denotes a part P from a ship S located at coordinates (X,Y).
{\small \begin{verbatimtab}[3]
domx(X) <- X in 1..10.                   
domy(Y) <- X in 1..10.

fol forall(S,Length,Part) :
    ship(S),ship_length(S,Length), Part in 1..Length 
    => ( exists(X,Y) : 
			domx(X),domy(Y), ship(S,Part,X,Y) ).
\end{verbatimtab} 
} 
\noindent Defining a ship by the locations of its parts introduces one
specific statement namely that the parts are connected to each other
and not separated by water or other boats.  The next statement ensures
also that the ships are either vertically or horizontally oriented.
{\small
\begin{verbatimtab}[3]
fol forall(S,P1,P2,X1,X2,Y1,Y2):
    ship(S,P1,X1,Y1), ship(S,P2,X2,Y2), P1 \= P2
    => ( X1 - X2 = P1 - P2, Y1=Y2 
		 ; Y1 - Y2 = P1 - P2, X1=X2).  
\end{verbatimtab}
} 
\noindent Another requirement is that two different ships do not touch each
other.  Translated to the above chosen representation of a ship, this
means that the distance between two parts belonging to different ships
is greater than 1.  
{\small
\begin{verbatimtab}[3]
fol forall(S1,S2,P1,P2,X1,X2,Y1,Y2):
    ship(S1,P1,X1,Y1), ship(S2,P2,X2,Y2), S1 \= S2
    => (abs(X1-X2) > 1 ; abs(Y1-Y2) > 1).
\end{verbatimtab}
}

The above statements describe the general knowledge about how ships are located.
In the context of a solitaire puzzle, the data specify for a subset of locations whether they contain water or a boat part.
We represent these data by a set of atomic fol axioms of the form: 
\bgroup \small 
\begin{verbatimtab}[3]
fol water(i,j).            fol boat(i,j)              
\end{verbatimtab} 
\egroup
\noindent where \texttt{water(i,j)} (\texttt{boat(i,j)}) means that on coordinate (i,j) there is water (a boat piece). 
As they exclude each other, we can define water as a location that is no occupied by a boat piece.
A location occupied by a boat piece is a location which is occupied by a part of a ship (as defined above).
Formally expressed
\bgroup\small \begin{verbatimtab}[3] 
water(X,Y) <- not boat(X,Y).  
boat(X,Y)  <- ship(S), ship_length(S,L), 
				  P in 1..L, ship(S,P,X,Y).
\end{verbatimtab} 
\egroup

Up to now there was no need for aggregates. 
But the puzzle gives also the number of ship parts located on a certain row or column.
As above we specify the data by a set of atomic fol axioms. 
\bgroup \small 
\begin{verbatimtab}[3]
fol row(i,n).         fol column(j,m).
\end{verbatimtab} 
\egroup
\noindent These two predicates are defined using the cardinality constraint as:
{ \small \begin{verbatimtab}[3]
row(I,N)    <- card(set([S,P], 
							(exists(Y): ship(S,P,I,Y)),N). 
column(J,M) <- card(set([S,P], 
							(exists(X): ship(S,P,X,J)),M).
\end{verbatimtab}
}

We used the above specification to solve a number of puzzles from the book \cite{battleship}.
The abductive solver was able to reduce the above specification to a finite domain constraint store.
This phase took about 1.5 second and is constant for all puzzles.
The time to find a solution of the constraint store varied: we obtained from 18 seconds for easy puzzles till 3 minutes for most difficult ones.
This means that most of this time is spend in enumerating candidate solutions by the CLP solver.
The time could be improved if some special search strategy as mentioned in the book, would be followed, but in our declarative approach it is not easy to specify a search strategy. This is an interesting topic for future research.

\subsection{Scheduling of maintenances}
The next experiment is based on a real life problem of a Belgian electricity provider.
The company has a network of power plants, distributed over different areas and each containing several power producing units.
These units need a fixed number of maintenances during the year and the problem is to schedule these maintenances so that the risk of power shortage (and hence, import from France) is as low as possible.
This approach extends earlier work described in \cite{Tractebel,Denecker97b}.

The fact that a maintenance {\tt M} lasts from week {\tt B} till week {\tt E}, is represented by the predicate \texttt{start(M,B,E)}. This is the only open predicate in the specification. Other predicates are either defined or are input data and are defined by a table. 
We will introduce now the constraints one by one.
\begin{itemize}
\item Maintenances (\texttt{maint(M)}) and their duration ({\tt duration(M,D)}) are given by a table. All maintenances must be scheduled, thus for each maintenance there exists an according \texttt{start} relation.
{ \small \begin{verbatimtab}[3]
fol forall(M) : maint(M)
    => exists(B,E,D): week(B), week(E), 
		 duration(M,D), (E = B + D -1), start(M,B,E).

week(W) <- W in 1..52.
\end{verbatimtab}
}
\item A table of \texttt{prohibited(U,Bp,Ep)} facts specify that maintenances \texttt{M} for unit {\tt U} are not allowed during the period {\tt [Bp,Ep]}:
{ \small \begin{verbatimtab}[3]
fol forall(U,Bp,Ep,M,B,E) :
    prohibited(U,Bp,Ep), maint_for_unit(M,U), 
	 start(M,B,E)
    => (E < Bp ; Ep < B).
\end{verbatimtab}
}
\item Some of the maintenances are not allowed to overlap.  The table of \texttt{non\_simult\_} \texttt{maint(M1,M2,Pre,Post)} facts describes this; \texttt{Post} and \texttt{Pre} represent the minimum
distance between the two maintenances.  
{ \small
\begin{verbatimtab}[3]
fol forall(M1,M2,Pre,Post,B1,E1,B2,E2) :
    non_simult_maint(M1,M2,Pre,Post), 
	 start(M1,B1,E1), start(M2,B2,E2)
    => (B2 > E1 + Post; B1 > E2 + Pre).
\end{verbatimtab}
}

\item Some maintenances should be done simultaneously, as defined by a table of \texttt{simult\_} \texttt{maint(M1,M2)} atoms. Two maintenances are simultaneous if the period of one is included in the period of another.
{ \small
\begin{verbatimtab}[3]
fol forall(M1,M2,B1,E1,B2,E2) :
    simult_maint(M1,M2), start(M1,B1,E1), 
	 start(M2,B2,E2)
    => ( (B1 =< B2, E2 =< E1) 
		 ; (B2 =< B1, E1 =< E2)).	
\end{verbatimtab}
}

\item  Different maintenances for the same unit should not overlap:  
{ \small
\begin{verbatimtab}[3]
fol forall(U,M1,M2,B1,E1,B2,E2) :
    unit(U), 
	 maint_for_unit(M1,U), maint_for_unit(M2,U), 
    M1 \= M2,start(M1,B1,E1), start(M2,B2,E2)
    => (E1 < B2; E2 < B1).
\end{verbatimtab}
}

\item For each week the number of the units in maintenance belonging to a plant \texttt{P} should be less than a maximal number \texttt{Max}. 
A given table of {\tt plant\_max(P,Max)} atoms defines for each plant the maximal number of units in maintenance simultaneously.  
{ \small
\begin{verbatimtab}[3]
fol forall(P,Max,We) :
    plant(P), plant_max(P,Max), week(We)
    => (exists(OnMaint): 
			card(set([U], (unit(U),unit_in_plant(U,P), 
								in_maint(U,We))), Onmaint),
          OnMaint =< Max ).
\end{verbatimtab}
}
We also define here a unit in maintenance, namely a unit is in maintenance during a certain week if there exists a maintenance \texttt{M} of this unit ongoing that week.
{ \small
\begin{verbatimtab}[3]
in_maint(U,W) <- exists(M,B,E) : 
					  maint_for_unit(M,U),start(M,B,E), 
					  B =<W, W=< E.
\end{verbatimtab}
}

\item The capacity of the units in maintenance belonging to a certain area should not exceed a given area maximum. To represent this, the summation aggregate is needed. A table of {\tt capacity(U,C)} describes for each unit its capacity.  
{ \small
\begin{verbatimtab}[3]
fol forall(A,Max,We,CapOnMaint):
    area(A),area_max(A,Max),week(We),
    sum(set([U], 
				(unit(U),in_area(U,A),in_maint(U,We))),
        lambda([Un],
					C where capacity(Un,C)), CapOnMaint)
    => 0 =< CapOnMaint, CapOnMaint =< Max.

in_area(U,A) <- unit_in_plant(U,P),
					 plant_in_area(P,A).
\end{verbatimtab}
}

\end{itemize}

The above specification describes the problem properly. Given input
data the solver comes up with a schedule for the
maintenances. However, the problem is to find an optimal solution that
keeps risk of power shortage  low. To do this, an optimality function
must be used.

This optimality function was proposed to us by the people of the
company. In the past, the company has kept track of the electricity
consumption during the year. These data can be used to compute an
estimate for the peak load consumption during each week. Given a
schedule, for each week one can compute the reserve capacity: the
difference between available capacity (i.e. the sum of capacities of
all units not in maintenance during this week) and the estimated
peak load. The optimization function is to maximize the minimal
reserve capacity over the year. 

\texttt{total\_capacity(T)} means that $T$ is the sum of all capacities of all units.
This is a constant value for the given problem.
 Peak loads are represented by a table of \texttt{peakload(Week,Load)} atoms.
 The predicate {\tt reserve(Week,R)} can be defined as follows: 
{ \small \begin{verbatimtab}[3] 
reserve(Week,R) <- exists(Load,T,InMaint) :
   peakload(Week,Load), total_capacity(T),
   sum(set([Unit], (unit(Unit), in_maint(Unit,Week))),
       lambda([U],C where capacity(U,C)), InMaint),
   R = T - Load - InMaint.
\end{verbatimtab}
}
Using this predicate, an optimal solution for the scheduling problem can be searched for if we add the following to our query
{ \small \begin{verbatimtab}[3]
minimum(set([R],(exists(W) : reserve(W,R)), M), 
maximize(M).
\end{verbatimtab} 
}
\noindent This means that we are interested in a solution in which the minimal reserve for one year is as high as possible.

Except for the representation of the optimization function, the above representation is very similar to the one used  in \cite{Denecker97b,Tractebel}. 
The actual problem, given by the company, consists of scheduling 56 maintenances for 46 units in one year.
The original system needed for this 24 hours to setup the CLP constraint store.  The bottleneck was the reduction of the aggregates. In the current implementation of SLDNFA the construction only takes 45 seconds.  
The huge difference comes from the fact that in the original system aggregates were implemented as large disjunctions, over which was backtracked until a consistent constraint store was found.
The current implementation will reduce the aggregates to a large finite domain constraint at once, and will not backtrack within it. 
It is the CLP solver which backtracks in the constructed constraint store; which is much faster than the abductive solver can do.
This difference also explains why the current implementation spends more time to find a good (optimal) solution (in 20 minutes we find a solution which is 94\% away from the optimal which is 2 or 3 times slower a the original one): the current constructed constraint store will contain all the disjunctions over which the original system had been backtracking to construct a small one.

A comparison with a pure CLP solution and the above solution shows clearly the tradeoff between declarative representation and a very fast solution.
The pure (optimized) CLP solution will setup its constraint store in several seconds (3 to 4 seconds), and find the same solution as the above specification within  2 minutes (compared to 20 minutes).
But on the other hand the CLP solution is a long program (400 lines) developed in some weeks of time  in which the constraints are hided within data structures,  where as the above representation is a simple declarative representation of 11 logical formulae, written down after some hours of discussion.

We want to stress the advantages of moving to an (even) more declarative representation than in the CLP solution. Development time, mentioned above, is just one of these advantages. Adaptability, extendibility and maintenance are others. It is our experience (and also reported in the experiments conducted with ACLP \cite{Kakas95a,Kakas99a}) that changes in the specification of the problem may result in several hours of work to adapt the CLP program. In the ID-logic representation, the same changes typically require a few minutes of work. The same distinction arises on the level of extendibility and maintenance. Taking these advantages into account, we believe that the reductions in speed mentioned above are a very good buy.

\section{Conclusion}
In a lot of real world problems statements involving aggregates are very common.
As such aggregates naturally show up in specifications.
This paper gives some preliminary results on the feasibility of using declarative specifications with aggregates to solve nontrivial computational problems. 

The extended abductive solver used in our experiments, is able to reduce the high level specification to a finite domain constraint store.
As can be expected, the generated constraint store tends to be more complex and less tuned to a specific problem than the ones generated by hand-written CLP programs. 
This disadvantage is covered by advantages as development time, adaptability, extendibility and easier maintenance.
However, at least in the context of the above experiments, reasonable efficiency could be obtained.
A way to improve the performance will be to optimize the generated constraint store.
Currently optimizations done in a hand-written CLP program aren't applied. 
We expect that an automated optimization will reduce the difference between both  substantially.
 
Further research needs to be done to get a better characterization of the class of problems the implementation can handle.
Together with a formal representation of the aggregate expression evaluation procedure we should be able to present a proof of correctness of the procedure for this class.
Another topic is how we can broaden this class without loosing to much efficiency.
Further we also expect that the introduction of tabling in the proof procedure will facilitate the treatment of aggregates considerably.

\section*{Acknowledgements}
Bert Van Nuffelen is supported by the GOA LP+ project at the K.U.Leuven.
Marc Denecker is supported by the FWO project "Representation and Reasoning in OLP-FOL".
We thank our colleagues from the DTAI-group at the K.U.Leuven for the discussions and useful comments. Especially the contribution of Danny De Schreye is appreciated for his help to clarify some points. 

\bibliography{/home/bertv/tekst/Papers/bertlib,/home/bertv/tekst/Papers/marclib}
\bibliographystyle{/home/bertv/tekst/LaTeX/aaai}

\end{document}